\documentclass{article}

\usepackage{microtype}
\usepackage{graphicx}
\usepackage{subfigure}
\usepackage{booktabs} %

\usepackage{hyperref}

\usepackage[accepted]{icml2023}

\usepackage{amsmath}
\usepackage{amssymb}
\usepackage{mathtools}
\usepackage{amsthm}

\usepackage[capitalize,noabbrev]{cleveref}

\usepackage{mathtools}
\usepackage{amsfonts}
\usepackage{empheq} %
\usepackage{cancel} %
\usepackage{custom_symbols}

\theoremstyle{plain}

\theoremstyle{definition}

\theoremstyle{remark}

\usepackage[textsize=tiny]{todonotes}

\icmltitlerunning{Exploiting Generalization in Offline Reinforcement Learning via Unseen State Augmentations}

\begin{document}

\twocolumn[
\icmltitle{Exploiting Generalization in Offline Reinforcement Learning \\via Unseen State Augmentations}

\icmlsetsymbol{equal}{*}

\begin{icmlauthorlist}
\icmlauthor{Nirbhay Modhe *}{gt,amzn}
\icmlauthor{Qiaozi Gao}{amzn}
\icmlauthor{Ashwin Kalyan}{ai2}
\icmlauthor{Dhruv Batra}{gt}
\icmlauthor{Govind Thattai}{amzn}
\icmlauthor{Gaurav S. Sukhatme}{amzn,usc}
\end{icmlauthorlist}

\icmlaffiliation{gt}{Georgia Institute of Technology}
\icmlaffiliation{amzn}{Amazon Alexa AI}
\icmlaffiliation{ai2}{Allen Institute for Artificial Intelligence}
\icmlaffiliation{usc}{University of Southern California}

\icmlcorrespondingauthor{Nirbhay Modhe}{nirbhaym@gatech.edu}

\icmlkeywords{Reinforcement Learning, Offline RL}

\vskip 0.3in
]

\printAffiliationsAndNotice{\icmlEqualContribution} %

\begin{abstract}
Offline reinforcement learning (RL) methods strike a balance between 
exploration and exploitation by conservative value estimation -- 
penalizing values of unseen states and actions.
Model-free methods penalize values at all unseen actions, 
while model-based methods are able to further exploit unseen states via model rollouts.
However, such methods are handicapped in their ability to find unseen states
far away from the available offline data due to two factors -- 
(a) very short rollout horizons in models due to cascading model errors, and
(b) model rollouts originating solely from states observed in offline data.
We relax the second assumption and present a novel unseen state augmentation
strategy to allow exploitation of unseen states where
the learned model and value estimates generalize. Our strategy finds unseen
states by value-informed perturbations of seen states followed by filtering out
states with epistemic uncertainty estimates too high (high error) or 
too low (too similar to seen data). We observe improved performance in several
offline RL tasks and find that our augmentation strategy consistently leads 
to overall lower average dataset Q-value estimates i.e. more conservative
Q-value estimates than a baseline.

Model-free methods penalize values at all unseen actions, while model-based methods are able to further exploit unseen states via model rollouts.
However, such methods are handicapped in their ability to find unseen states far away from the available offline data due to two factors -- (a) very short rollout horizons in models due to cascading model errors, and (b) model rollouts originating solely from states observed in offline data.
We relax the second assumption and present a novel unseen state augmentation strategy to allow exploitation of unseen states where the learned model and value estimates generalize. 
Our strategy finds unseen states by value-informed perturbations of seen states followed by filtering out states with epistemic uncertainty estimates too high (high error) or too low (too similar to seen data). 
We observe improved performance in several offline RL tasks and find that our augmentation strategy consistently leads to overall lower average dataset Q-value estimates i.e. more conservative Q-value estimates than a baseline.

\end{abstract}

\section{Introduction}
\begin{figure}[t]
\centering
    \resizebox{0.95\linewidth}{!}{%
        \includegraphics[width=0.98\linewidth]{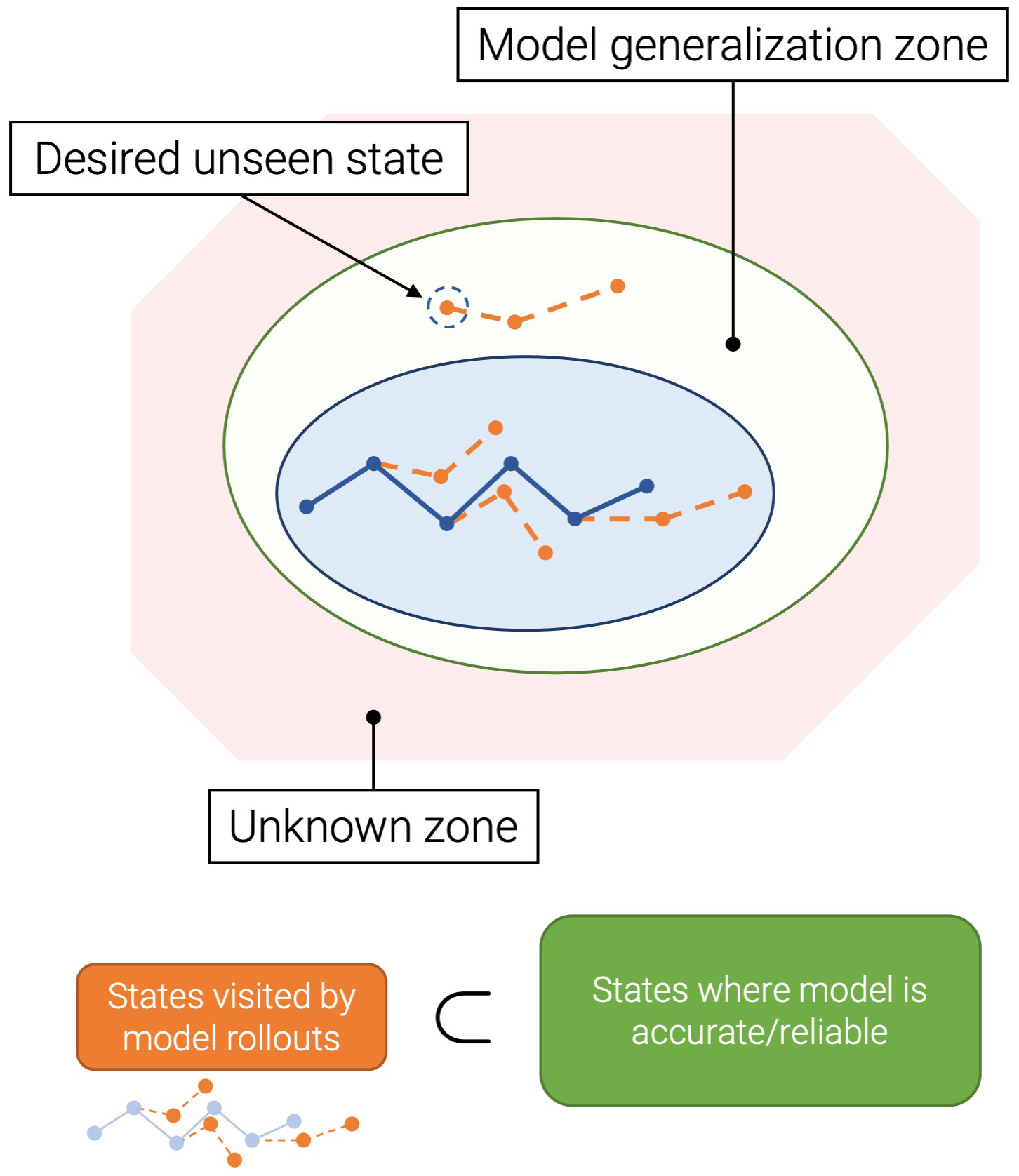}
    }
\caption{%
We hypothesize that there exists a `green zone' of model generalization
that is not covered by the extent of model rollouts with short horizons 
(innermost blue zone). We propose a strategy to find such states and show
that this leads to improved offline RL performance and 
lower conservative Q-value estimates.
}
\label{fig:adq_fig}
\end{figure}

Offline reinforcement learning (offline RL, batch RL) \citep{levine2020offline,lange2012batch,fujimoto2019off} is a uniquely challenging problem due to the
offline nature of data being used to learn policies, without any new environment interactions.
It has also gained relevance due to the increasingly large amounts of 
offline data available for gleaning useful behaviors without the need for training from scratch. 
Offline RL has been applied broadly -- 
in robotics, machinery and healthcare \cite{levine2020offline,wang2018supervised,kalashnikov2021mt,zhan2022deepthermal}, 
where exploration is costly and often dangerous.

Offline RL algorithms have an inherent risk trade-off -- better performance beyond that of the
observed behaviors in offline data comes with higher uncertainty in value estimates from states and actions far away 
from seen data. Traditional offline RL methods realize this trade-off by either penalizing deviation from observed behavior 
\cite{fujimoto2018off,yu2020mopo} or by conservatively estimating values for unseen actions 
such as in Conservative Q-Learning (CQL \cite{kumar2020cql}).
While model-free methods have prescribed ways for dealing with unseen actions, model-based offline RL methods \cite{yu2021combo,kidambi2020morel,yu2020mopo} utilize a model of the transition dynamics and reward in order to leverage generalization to unseen states in addition to unseen actions. 
However, model-based offline RL methods are not able to meet such expectations 
of generalization to unseen state due to a combination of two limiting factors -- 
(i) value estimates via model-based value expansion (MVE \cite{revisiting_mve,mbve_for_mfrl}) 
are made starting from seen states alone, ignoring those unseen states
where learned estimators may have low error, and 
(ii) the rollout horizon for model-based value expansion is typically very short
due to compounding errors in long horizon future predictions.

We hypothesize that expanding model-based value estimation to carefully selected unseen states
will improve state-action value estimation at unseen states while maintaining the conservative
bias for unseen quantities (states as well as actions) as prescribed in prior works such 
as CQL \cite{kumar2020cql} and COMBO \cite{yu2021combo}. We design an unseen state generation
strategy with a \emph{propose} and \emph{filter} strategy -- 
(i) candidate unseen states proposals are generated via value-informed perturbations to seen states,
(ii) candidates that have too high estimated uncertainty (likely to have high 
model and value estimation error) or too low epistemic uncertainty (likely to be too similar to seen states and 
less informative) are discarded. 

We present empirical evidence of improved performance on
the D4RL benchmark \cite{fu2020d4rl} and several analyses of the mechanism of action of our method.
We show the reliability of our epistemic uncertainty estimation is predicting true model error, demonstrating
that it is more suitable than aleatoric uncertainty estimators that have been shown to have poor correlation
with true model error \cite{yu2021combo}. 
We study the importance of our value-informed perturbations
by (i) measuring the informativeness of value-informed state perturbation as opposed
to perturbing states in random directions and (ii) ablating the direction of value-informed perturbations.
Finally, we present an analysis of average dataset Q-value, 
the key metric used in prior work (COMBO \cite{yu2021combo}) for offline hyperparameter selection, and demonstrate
that our perturb and filter strategy leads to significantly lower average dataset Q-value estimates
on top of COMBO (used as a baseline) -- hinting that the mechanism of action for improved performance
may be in decreased Q-value estimates in the face of conservative penalties.

\section{Related Work}
Offline or Batch Reinforcement Learning 
is the problem of learning optimal quantities (values or policies)
from a fixed set of experiences or data, with its roots in fitted value and Q iteration 
\cite{gordon1995stable, riedmiller2005neural} and has recently gained popularity due to the
success of powerful function approximators such as deep neural
networks in learning policies, values, rewards and dynamics 
\cite{levine2020offline,yu2020mopo,yu2021combo,kidambi2020morel}.
\nmf{Offline RL is distinct from imitation learning as the latter assumes expert demonstrations
as offline data without reward labels, whereas offline RL assumes known reward labels, 
allowing for use of sub-optimal or non-expert data.}
Offline RL datasets such as D4RL \cite{fu2020d4rl} and NeoRL \cite{qin2021neorl} are prominent
benchmarks for evaluating and comparing offline RL methods. We use environments
from D4RL in this chapter for evaluating our method.

\textbf{Unlabeled Data in Offline RL:} Offline RL assumes that the fixed dataset
consists of tuples of state, action, next state and reward. However,
unlabeled data i.e. without reward labels is often plentiful and an the question
of how to leverage such data to improve Offline RL with a typically 
smaller labeled dataset is an important question. Setting the reward of 
unlabeled data to zero has been shown to be a simple and effective strategy 
to leverage this data \cite{yu2022unlabeled_offrl}. Intuitively, this is reasonable
to expect as Q-values from reward labeled states will propagate to unlabeled states
during learning of a Q-function with temporal difference.

\textbf{Conservative Q-Learning (CQL):} A number of works have used or adapted the method
of learning conservative Q estimates by penalizing the estimated value of unseen actions
while pushing up value of seen actions in the offline dataset \cite{kumar2020cql}.
\citet{yu2021combo} is a model-based approach that combines CQL \cite{kumar2020cql}
with model-based reinforcement learning by additionally penalizing the estimated
values of states actions in model rollouts.

\textbf{Model-based Offline RL:} COMBO \cite{yu2021combo}, MOREL \cite{kidambi2020morel} 
are model-based offline RL approaches that use the approximate MDP induced by a learned dynamics
model network with a conservative objective. As previously mentioned, 
\citet{yu2021combo} uses CQL by penalizing estimated value of state-actions visited
during model rollouts stating from seen states, while \citet{kidambi2020morel}
defines their induced MDP in a conservative manner by stopping any model rollouts
that fall into states with high uncertainty as measured by model ensemble disagreement.
COMBO \citet{yu2021combo} expressly avoids uncertainty estimation citing its
inaccuracy in some settings. 
There are two key differences w.r.t COMBO --
(i) we use epistemic uncertainty estimation and show its correlation with true model error,
whereas COMBO demonstrated the poor correlation of aleatoric uncertainty estimation with 
true model error; (ii) 
we do not re-introduce uncertainty in the manner COMBO recommended to avoid -- 
we use uncertainty to filter unseen states used for starting model rollouts 
as opposed to using it as a reward penalty (as in MOPO \cite{yu2020mopo}).

\textbf{Adversarial Unseen State Augmentation:} \citet{zhang2021adversarial} is a model-free
online RL approach that proposes adversarial states as data augmentation for policy updates
via multiple gradient steps in state space. This is similar to our approach that proposes
unseen states via multiple gradient updates in the state space of the Q-value estimate.
Other than the difference of objective for computing the gradient, there are also two 
additional differences between \citet{zhang2021adversarial} and our approach -- (1)
we use both the positive and negative gradient direction for proposing unseen states
and find that both are jointly more useful than one direction alone, whereas
\citet{zhang2021adversarial} is purely adversarial (minimizing their objective with gradient descent), 
and (2) \citet{zhang2021adversarial}
keep the sampled action fixed while minimizing their adversarial objective while we
use the differentiability of the parameterized policy to sample new actions.

\section{Preliminaries}

In this section, we briefly touch upon Offline Reinforcement Learning (RL) and
the baseline COMBO \cite{yu2021combo} that we build upon and compare with
in our analysis.
\begin{figure*}[t]
\centering
    \resizebox{1.0\textwidth}{!}{%
        \includegraphics[width=0.98\textwidth]{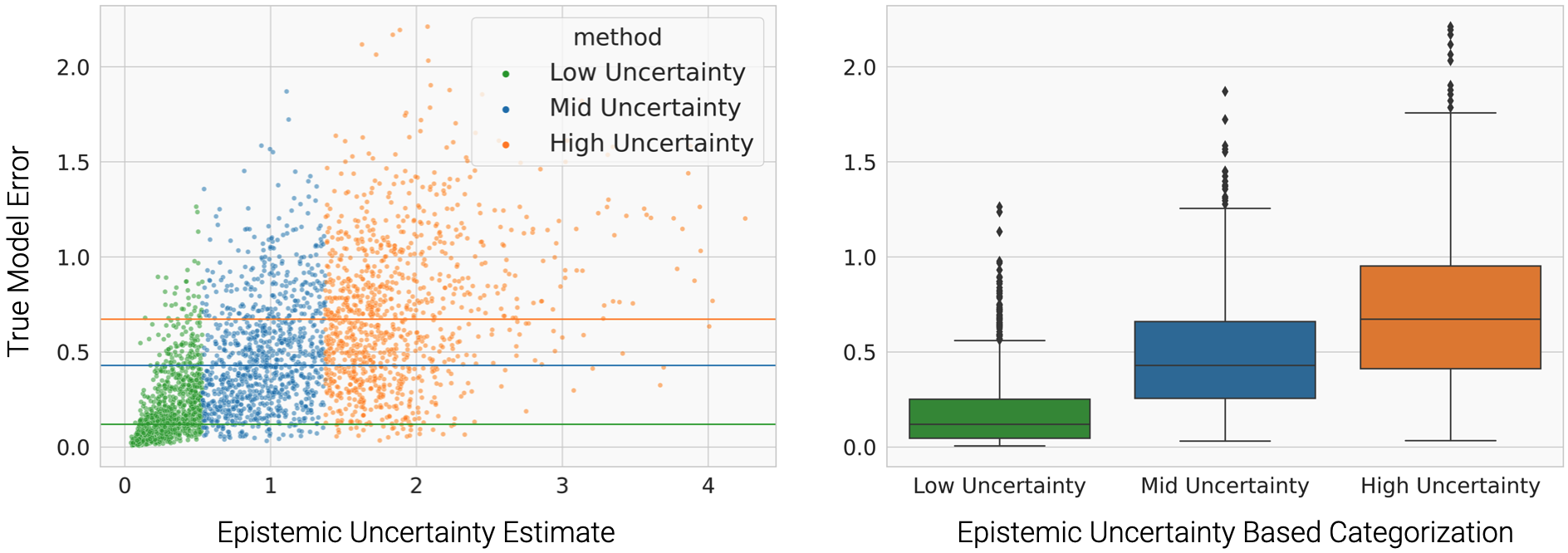}
    }
\caption{%
(Left) Relationship between epistemic uncertainty estimated using ensemble disagreement
and true model error i.e. the mean absolute difference between 
model next state and reward predictions and true quantities, for the
Adroit Pen manipulation task and Human demonstrations dataset.
Each point corresponds to a state obtained by model rollouts from a perturbed
state in \Cref{alg:value_aware_augment}.
Horizontal lines measure medians for respective uncertainty category.
(Right) Box plot of true model error aggregated for each uncertainty
category. The two boundaries for the categories are the 0.25 and 0.75
quantile uncertainty estimates on the seen states in the offline dataset.
The states filtered into the Mid Uncertainty class are used
for unseen state augmentation.
}
\label{fig:epistemic_unc}
\end{figure*}

\textbf{Offline RL.}
In Offline Reinforcement Learning, a fixed dataset of interactions $\calD:={(s, a, r, \sprime)}$
is provided for finding the best possible policy without any environment interactions 
(i.e. without ``online'' interactions). Since we only deal with simulated environments,
the policy found by an algorithm is evaluated with online evaluation.

\textbf{Model-based Offline RL.}
COMBO \cite{yu2021combo} is a model-based offline RL algorithm that extends the concept
of conservative Q-learning in CQL \cite{kumar2020cql} to model-generated predictions.
\Cref{alg:mboffrl_combo} depicts the model-based RL algorithm used by COMBO. First,
an initial model fitting stage is executed where the reward and dynamics 
model is trained with 
maximum likelihood estimation (MLE) on 
the offline data. 
Second, policy and critic learning
happens over a fixed number of training epochs where the data used for 
policy training is obtained by means of \emph{branched rollouts}. Branched
rollouts are inherited from the model-based algorithm MOPO \cite{yu2020mopo}
(which COMBO uses as a foundation),
where a batch of initial states are sampling from the offline dataset and
used to generate model-rollouts up to a fixed horizon $H$ using actions
sampled from a rollout policy $\mu(\cdot\mid s)$ (typically the same
as $\pi$ or sometimes a uniform over actions policy) and next states
sampled from the learned dynamics model (almost always Gaussian with mean
and variance predictions). Then, the policy and critic are updated
using this rollout data.

The conservative Q-value update used by COMBO is as follows. Here,
$d_{f}$ is an interpolation defined by \cite{yu2021combo}
as $d_f(s, a):=f d(s, a)+(1-f) d^{\mu}_{\rMphi}(s, a)$; where
$d$ is the empirical sampling distribution from the offline dataset $\calD$.

\begin{align}
\nonumber
\Qhat^{k+1} &\leftarrow 
\argmin_Q 
    \calL_{\textrm{TD}} + \beta \cdot \calL_{\textrm{CQL}}
\\
\nonumber
\calL_{\textrm{TD}}
&:= 
\mEE_{(s, a, \sprime) \sim d_{f}}
\Bigg[
\bigg(
    \bigg(
        r(s, a)
        \nonumber\\&\phantom{AAA}\nonumber
        +\gamma \mEE_{
            a^{\prime} \sim \pihat^k
            \left(
                a^{\prime} \mid \sprime
            \right)
        }
        \left[
            \Qhat^k
            \left(
                \sprime, a^{\prime}
            \right)
        \right]
    \bigg)
    -Q(s, a)
\bigg)^2
\Bigg]
\\
\calL_{\textrm{CQL}}
&:= \mEE_{s\sim d_{\rMphi,\pi}, a\sim \pi}\left[
    Q(s, a)
\right]
- \mEE_{(s, a)\sim \calD}\left[
    Q(s, a)
\right]
\label{eqn:combo_q_update}
\end{align}

The right hand term in \cref{eqn:combo_q_update} is further used for
offline tuning of hyperparameters in COMBO -- a trait that we inherit
when using COMBO as a baseline.
Conservative Q-value
estimation is applied to all rollout data (i.e. Q-values are pushed
down for unseen states predicted by the model). 
This implies that while Q-value
propagation does occur for unseen states found by model rollouts,
they are down-weighted proportional to the coefficient $\beta$ in the
above equation \eqref{eqn:combo_q_update}. This implies that
when we do find unseen states (in \cref{sec:approach}) 
in regions not visited by branched rollouts,
they will also bear the same conservative penalty.

\begin{algorithm}[t]
\caption{Model-based Offline RL with COMBO \cite{yu2021combo}}
\label{alg:mboffrl_combo}
\begin{algorithmic}
\STATE {\bfseries Input:}{Offline dataset $\calD$, parametrized policy and critic $\pi_{\theta}$, $Q_{\theta}$}
\STATE {\bfseries Input:}{Parametrized dynamics model $\calP_{\rphi}$}
\STATE {\bfseries Input:}{Rollout policy $\mu(\cdot\mid s)$, horizon $H\in\mathbb{N}$}
\STATE Split $D$ into train, val splits and optimize $\calP_{\rphi}$ until convergence on val set;
\FOR{$i=1$ to $n_{\textrm{epochs}}$}
    \STATE $\calD_0 \leftarrow$ Sample a batch of start states from $\calD$;
    \STATE $\calDhat \leftarrow$ Collect model rollouts with $\mu,\calP_{\rphi}$ up to $H$ starting from states in $\calD_0$;
    \STATE Fit $Q_{\theta}$ with a conservative objective using both $\calD$ and $\calDhat$;
    \STATE Update policy $\pi_{\theta}$ using estimated $Q_{\theta}$ values on $\calDhat$;
\ENDFOR
\end{algorithmic}
\end{algorithm}

\section{Approach}\label{sec:approach}
In this section,
we first motivate our method by highlighting the 
limitations of using branched model rollouts
along with short model prediction horizons.
This raises the question of how to find unseen states for model 
rollouts in a manner that is accurate (unseen
states have low model error during rollouts) 
and useful (leads to better downstream task performance).
To achieve this,
we introduce a simple propose and filter
strategy for finding unseen states -- a proposal stage where a 
seen state is perturbed and a filter stage where states having
estimated uncertainty too low or too high are discarded.
We then propose two perturbation functions for obtaining unseen
state proposals -- one that perturbs randomly uniformly in any
direction and a second that perturbs along the positive and negative
direction of the Q-value gradient w.r.t states.

\textbf{Limitations of Short Horizon Branched Rollouts.}
In offline-RL, branched model rollouts utilize every state in the 
seen dataset for performing rollouts i.e. sequential 
model next state predictions given a behavior policy, up to horizon $H$.
Such unseen state predictions populate the rollout buffer and along with seen states,
they are used for Q-learning
using temporal difference and conservative objectives (CQL \citet{kumar2020cql}).
Ideally, to fully exploit the learned model, all states where the model
has low uncertainty should be included in the rollout buffer for Q-learning.
However, this may not be the case as there is no guarantee that
the states visited by $H$ horizon model rollouts will visit all possible
states where the model has low uncertainty i.e. states where the learned
dynamics and reward model generalizes.
We will show in our experiments that exploiting such unseen states consistently 
leads to lower Q-value estimates.

\textbf{Start State Augmentation.}
In order to maximize coverage of states where the learned dynamics and reward model
generalize to, as measured by estimated uncertainty,
we propose an augmentation strategy that uses a mixture of seen and unseen states as
starting points to perform model rollouts. Given a batch of $\nstart$ states 
sampled from the offline dataset, we control the fraction of this batch to replace
with unseen states as $\faug \in [0, 1]$. 
We find that $\faug = 0.5$ works well in practice. Note that $\faug=0$ reduces to a 
no augmentation baseline.

\textbf{Perturb and Filter with Q-gradients.}
\Cref{alg:value_aware_augment} details the subroutine, referred to as \PNF,
that takes as input a batch of states sampled from the offline dataset and produces
a set of potentially unseen states via a perturb and filter strategy.
First, a set of candidate state proposals are generated by taking
$\nsteps$ gradient steps using the Q-value gradient w.r.t state input,
with a step size uniformly sampled from $(-\delmax,\delmax)$ i.e.
with positive as well as negative directions, and fixed across gradient
steps. Next, uncertainty cut-off points $\ulow, \uhigh$, computed
using the $0.25$ and $0.75$ quantile uncertainty values of seen data, 
are used for filtering out states that lie outside of the range $(\ulow, \uhigh)$.
This process is repeated until the desired number of augmented states $\naug$
have been obtained. The while loop typically runs for just one iteration
in most of our hyperparameter choices.

\textbf{Uncertainty Estimator.} We tested 3 types of uncertainty estimators:
(1) ensemble disagreement via max 2-norm deviation from mean across ensemble 
(used in MOPO implementation by \cite{qin2021neorl}),
(2) maximum 2-norm of standard deviation across ensemble 
from (used by MOPO \cite{yu2020mopo}) and (3) standard deviation of mean predictions
over ensemble. Here, the first and third estimators approximate 
epistemic uncertainty while
the second approximates aleatoric uncertainty. We found (1) and (3) to overall 
work well in practise and we select (1) as our choice of uncertainty estimator 
for all of our experiments.
We exclude MOREL's discrepancy based uncertainty \cite{kidambi2020morel}, as we found it to be 
similar to (1) in form and practice. \cref{fig:epistemic_unc} (left) shows the correlation between
epistemic uncertainty (using (1)) and true model error for the 
Adroit Pen manipulation task with the Human demonstrations dataset. 
\cref{fig:epistemic_unc} (right) demonstrates our strategy for filtering states -- defining
a minimum and maximum threshold based on the 0.25 and 0.75 quantile uncertainty values
computed on the seen states in the offline dataset.

\textbf{Ablations.} In addition to using the Q-value gradient for 
generating state proposals, we also test an ablation of our method named
\PNFRand, that substitutes the Q-value gradient in 
\Cref{alg:value_aware_augment}, \Cref{line:q_grad_dir}
with a unit vector sampled uniformly randomly on a hypersphere.
In order to prevent random walks, we fix $\nsteps$ to $1$ for this ablation
so that the augmented states are solely a result of a single perturbation
per direction.
We find that the Q-gradient produces more informative
state proposals after filtering than this \PNFRand, 
as measured by final policy performance
on tested offline RL benchmarks.

\begin{algorithm}[t]
\caption{\PNF: Value-aware unseen state augmentation}
\label{alg:value_aware_augment}
\begin{algorithmic}
\STATE {\bfseries Input:}{Input state batch $\Dbatch\subseteq\calD$, 
parametrized Q-network $Q_{\theta}$, policy network $\pi_{\theta}$}
\STATE {\bfseries Input:}{State-dependent uncertainty estimation function $\uncf:\calS\rightarrow\RR$}
\STATE {\bfseries Input:}{Number of steps of perturbation $\nsteps$, maximum step size $\delmax$}
\STATE {\bfseries Input:}{Desired number of augmented states $\naug$ to fill $\Daug$}
\STATE Initialize $\Daug \leftarrow \emptyset$ and $U_0 := \{ \uncf(s): s\in \Dbatch \}$;
\STATE Initialize $\ulow$, $\uhigh$ as $0.25$-quantile and $0.75$-quantile of $U_0$ respectively;
\WHILE{$\abs{\Daug} < \naug$}
    \STATE $\Dbar_{0} \leftarrow \{s : (s, a, r, \sprime)\in \Dbatch\}$;
    \STATE Sample $\eta_{s} \sim U(-\delmax, \delmax) \forall s\in\Dbar_0$;
    \FOR{$i$ from $1$ to $\nsteps$}
        \STATE $\Dbar_i \leftarrow \{ \sbar : \sbar \leftarrow s 
            + \eta_{s} \cdot \nabla_{s} Q_{\theta}(s, \pi_{\theta}(s)) \}$;
            \label{line:q_grad_dir}
    \ENDFOR
    \STATE $\Dbar_{\textrm{all}} \leftarrow \bigcup_{i=1}^{\nsteps} \Dbar_i $;
    \STATE $\Dbar_{\textrm{filtered}} \leftarrow 
    \{s : \uncf(s) > \ulow \textrm{ and } \uncf(s) < \uhigh\ \forall \s \in \Dbar_{\textrm{all}} \} $;
    \STATE $\Daug \leftarrow \Daug \bigcup \Dbar_{\textrm{filtered}}$ ;
    \STATE If $\abs{\Daug} > \naug$, subsample uniformly to keep $\naug$ elements;
\ENDWHILE
\end{algorithmic}
\end{algorithm}

\begin{algorithm}[t]
\begin{algorithmic}
\caption{Model-based Offline RL with State Augmentation}
\label{alg:mboffrl_ours}
\STATE {\bfseries Input:}{Offline dataset $\calD$, parametrized policy and critic $\pi_{\theta}$, $Q_{\theta}$}
\STATE {\bfseries Input:}{Parametrized dynamics model $\calP_{\rphi}$}
\STATE {\bfseries Input:}{Rollout policy $\mu(\cdot\mid s)$, horizon $H\in\mathbb{N}$}
\STATE {\bfseries Input:}{Fraction of start state batch to augment $\faug\in[0,1]$}
\STATE Split $D$ into train, val splits and optimize $\calP_{\rphi}$ until convergence on val set;
\FOR{$i=1$ to $n_{\textrm{epochs}}$}
    \STATE $\Dstart \leftarrow$ Sample a batch of $\nstart$ start states from $\calD$;
    \STATE $\Dstart \leftarrow$ Augment {$\naug:=\floor{\faug \cdot \nstart}$} states using \Cref{alg:value_aware_augment};
    \STATE $\calDhat \leftarrow$ Collect model rollouts with $\mu,\calP_{\rphi}$ up to $H$ starting from states in $\Dstart$;
    \STATE Fit $Q_{\theta}$ with a conservative objective using both $\calD$ and $\calDhat$;
    \STATE Update policy $\pi_{\theta}$ using estimated $Q_{\theta}$ values on $\calDhat$;
\ENDFOR
\end{algorithmic}
\end{algorithm}

\section{Experiments}
In this section, we perform empirical analyses to answer the following
questions: 
(1) Does our \tPNF method lead to improvemenet in offline RL
performance over a baseline? 
(2) Does the Q-gradient direction matter
for perturbation, when compared to performance of a PnF-Random baseline?
(3) Does the sign of the perturbation (i.e. ascending or descending
the Q-gradient) matter for performance? 
and 
(4) What are the distance characteristics of the unseen states found
with PnF augmentation as opposed to the unseen states visited by model
rollouts in the baseline?

\subsection{Performance on D4RL Environments and Datasets}
We use COMBO \cite{yu2021combo} as a foundation on top of which we
implement our method \PNF. We use the open source 
PyTorch implementation of COMBO by \cite{qin2021neorl}.
Our method requires specification of the following additional hyperparameters.
\begin{enumerate}
    \item $\nsteps$: Number of gradient descent or ascent steps in 
    \Cref{alg:value_aware_augment}. We use the range $\{1, 2, 4, 8\}$
    to pick the best value.
    \item $\delmax$: Maximum value of step size $\eta_{s}$ in 
    \Cref{alg:value_aware_augment}. We use the (logarithmic) range 
    $\{1.0E-5, 5.0E-5, 1.0E-4, 5.0E-4, 1.0E-3, 5.0E-3, 1.0E-2, 5.0E-2, 1.0E-1, 0.5\}$
    to pick the best value.
    \item $\faug$: Fraction of batch of start states to augment in 
    \Cref{alg:mboffrl_ours}. We use the range $\{0.5, 0.9, 1.0\}$
    to pick the value, with the mimunum fraction of $0.5$ guaranteeing that
    at least half of the batch is augmented.
\end{enumerate}

We first compare PnF-Qrad performance to COMBO on several offline RL 
environments from the D4RL benchmark \cite{fu2020d4rl}.
We refer to our method as \tPNF to emphasize that
COMBO is the base algorithm on top of which we perform 
unseen state augmentation. Similarly, we refer to our method with
random perturbation directions as \tPNFRand.

\begin{figure}[t]
\centering
    \resizebox{1.0\linewidth}{!}{%
        \includegraphics[width=0.98\textwidth]{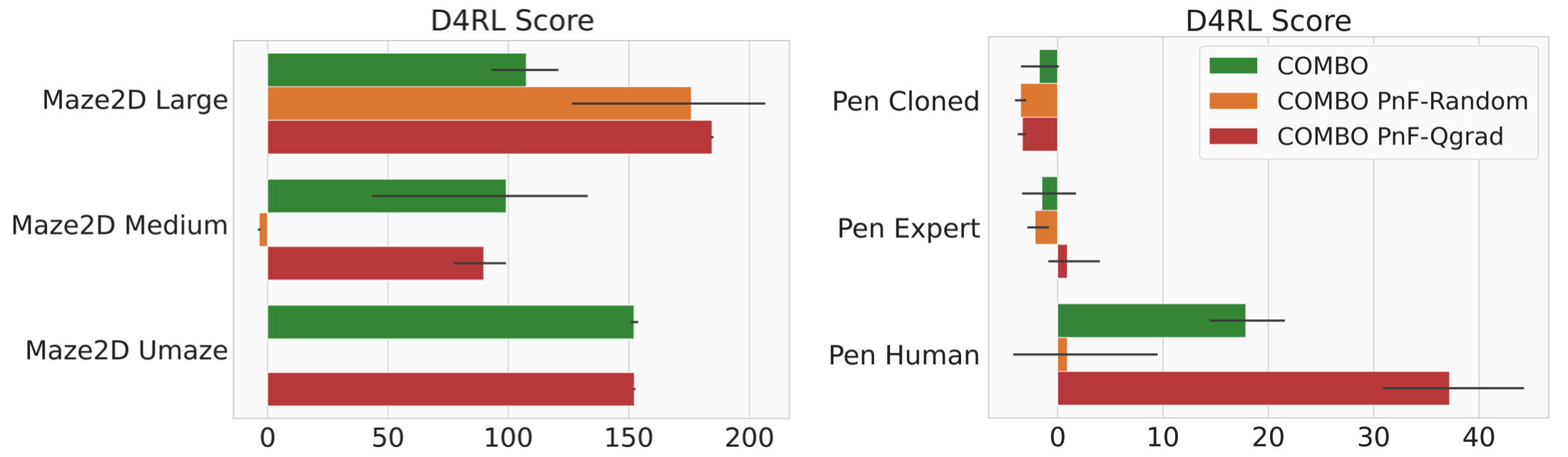}
    }
\caption{%
Evaluation on two sets of D4RL offline datasets, the Maze2D environment
with varying maze sizes (left) 
and Adroit Dextrous Hand Manipulation - Pen Environment with 
varying dataset types (right). 
The Maze2D dataset contain a little under $4$ million time steps.
The \texttt{Pen-v1 Expert} and \texttt{Pen-v1 Cloned}
datasets contain close to $\sim495000$ time steps, whereas 
the \texttt{Pen-v1 Human} dataset contains $4950$ time steps.
Hyperparameters are tuned on \texttt{Maze2D-v1 Medium} for the 
Maze2D tasks and \texttt{Pen-v1 Human} for the Adroit Pen tasks.
}
\label{fig:adroit_maze_results}
\end{figure}

In D4RL, a task (e.g. \texttt{Pen-v1 Human}) 
is specified by a choice of environment
(e.g. Adroit Pen environment) and a choice of dataset 
(e.g. Human demonstrations).
\Cref{fig:adroit_maze_results} evaluates \tCOMBO, 
\tPNF and \tPNFRand on two
types of D4RL tasks -- the Maze2D family of tasks that 
have three different environments corresponding to varying
maze grid sizes (\texttt{Umaze}, \texttt{Medium}, \texttt{Large}).
We find that on the \texttt{Pen-v1 Human} task consisting of
4950 time steps of human demonstrations, \tPNF
significantly outperforms baselines. However,
on the \texttt{Pen-v1 Cloned} and \texttt{Pen-v1 Expert} tasks,
we find that all methods including baselines perform poorly.
In the Maze2D tasks, we find an improvement over \tCOMBO
for both \tPNF and \tPNFRand on the \texttt{Maze2D-v1 Large} task.
\tPNFRand performs poorly on the \texttt{Maze2D-v1 Medium} and 
\texttt{Maze2D-v1 Umaze} tasks where \tCOMBO and \tPNF
perform comparably.

\textbf{Hyperparameter Sensitivity}
In \Cref{tab:mujoco_results}, we compare \tPNF, \tCOMBO (our implementation
using \cite{qin2021neorl} and the reported performance by Yu et al. in 
\cite{yu2021combo}. Due to computational resource constraints, we tune
hyperparameters on a single task -- the \texttt{Walker2D-v2 Medium-Replay}
task, while evaluating on all other tasks. We find that, while we do
observe statistically significant improvement in performance on the 
same environment where hyperparameter values were tuned, we don't see
a complete replication of this trend when transferring hyperameters to other
tasks. We suspect that this sensitivity is a result of the varying
distribution and distance between states in each dataset and environment,
as these will affect the perturbation magnitude in our method \PNF
(e.g. value of $\delmax, \nsteps$).

\begin{figure}[h!]
\centering
    \resizebox{1.0\linewidth}{!}{%
        \includegraphics[width=0.98\textwidth]{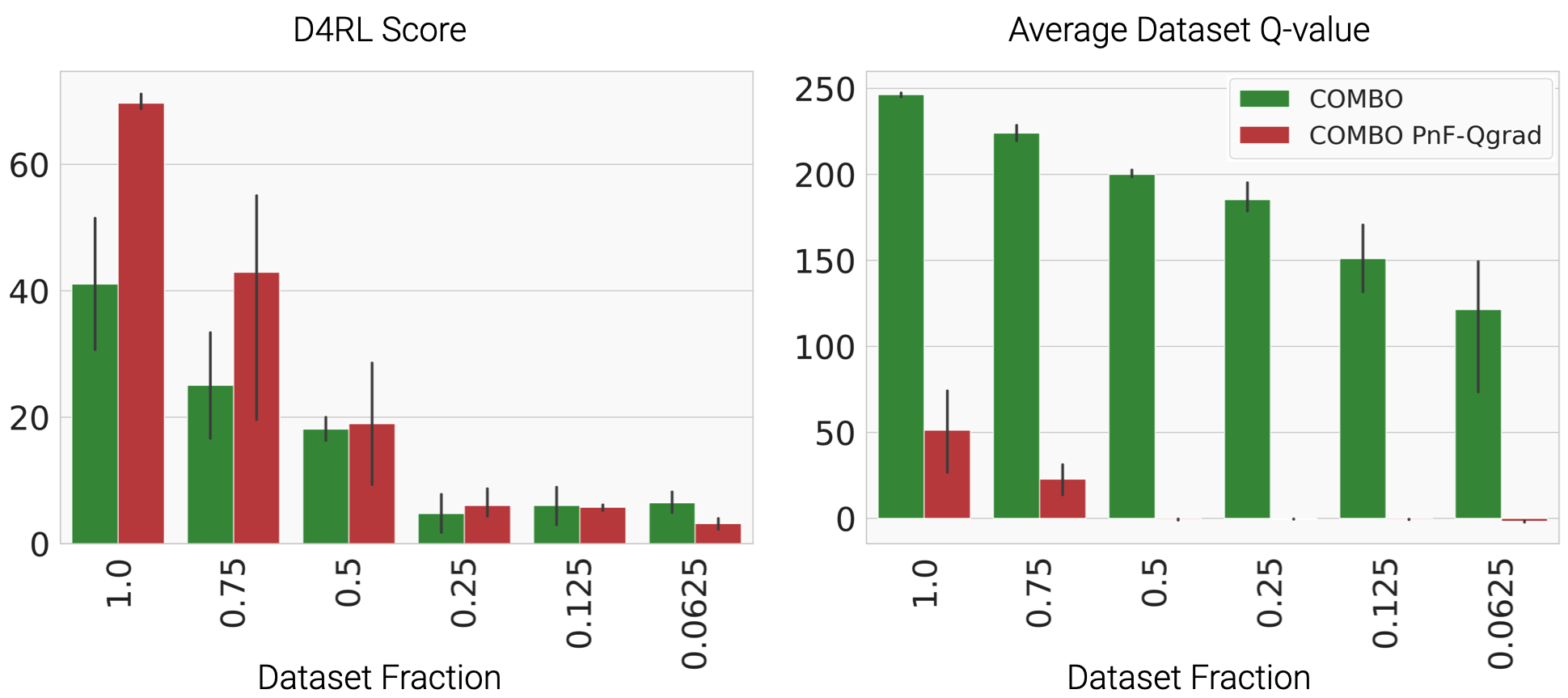}
    }
\caption{%
D4RL score (left) and average dataset Q-value (right) for varying dataset size fractions
for the \texttt{Walker2D-v2 Medium-Replay} task. Each dataset fraction corresponds to a
contiguous subarray taken from the front of the original dataset. We observe significantly
lower average dataset Q-values but comparable D4RL scores for \texttt{COMBO \PNF} 
versus \texttt{COMBO} as the dataset size reduces.
}
\label{fig:dset_frac}
\end{figure}

\textbf{Average Dataset Q-value}
In \Cref{fig:adq_fig}, we measure the average dataset Q-value, a quantity
important for not only selecting hyperparameters in an offline manner,
but one which has shown to indicate online evaluation performance. COMBO
\cite{yu2021combo} has shown lower average dataset Q-values lead to more 
conservative Q-value estimates i.e. preventing any overestimation of Q-values,
and higher online evaluation performance (D4RL score). We find that throughout
the 500 epochs of policy training, average dataset Q-value is significantly
lower for \texttt{COMBO \PNF} in comparison to \texttt{COMBO}. Further,
in \Cref{fig:dset_frac}, we see that this trend holds when reducing the
size of the dataset, despite the performance of the two being comparable
at lower dataset sizes. This suggests that conservative Q-value estimation
is an important mechanism of action for our proposed method in contributing
to better online evaluation performance.

\begin{table*}[h]
\caption{
D4RL Score mean and standard error over 6 random seeds for (left to right) 
COMBO (reported in \protect\cite{yu2021combo}), 
the implementation of COMBO by \protect\cite{qin2021neorl}
that we use and our proposed \PNF method.
\PNF requires tuning of the hyperparameters $\delmax, \nsteps, \faug$ which
are sensitive to both datasets and environments. We tune these hyperparameters on the 
\texttt{Walker2D-v2} \texttt{Medium-Replay} dataset, according to the offline tuning 
guidelines from \protect\citep{yu2021combo,kumar2021workflow}, 
and report online evaluation performance of the selected
hyperparameters across all other environments and datasets. 
$*$ indicates tuned
on \texttt{Walker2D-v2} \texttt{Medium-Replay}.
}
\begin{center}
\includegraphics[width=0.65\linewidth]{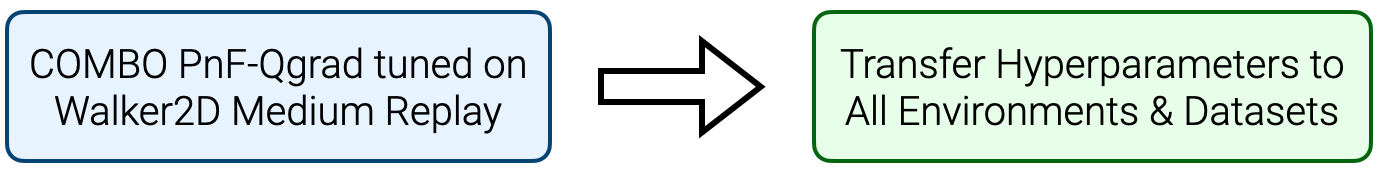}\\
\hspace{1cm}
\resizebox{0.9\linewidth}{!}{%
\begin{tabular}{ccccc}
\hline
 \textbf{Dataset type}   & \textbf{Environment}   & \textbf{COMBO (Yu et al.)}   & \textbf{COMBO}   & \textbf{COMBO PnF-Qgrad*}   \\
\hline
 random                  & halfcheetah            & $38.8 \pm 3.7$              & $14.5 \pm 6.4$   & $-1.5 \pm 0.5$              \\
 random                  & hopper                 & $17.9 \pm 1.4$              & $ 4.7 \pm 1.8$   & $ 4.6 \pm 2.6$              \\
 random                  & walker2d               & $ 7.0 \pm 3.6$              & $ 7.6 \pm 1.4$   & $ 5.1 \pm 1.8$              \\
 medium                  & halfcheetah            & $54.2 \pm 1.5$              & $62.6 \pm 1.3$   & $41.2 \pm 4.5$              \\
 medium                  & hopper                 & $97.2 \pm 2.2$              & $61.0 \pm 0.4$   & $51.6 \pm 15.5$             \\
 medium                  & walker2d               & $81.9 \pm 2.8$              & $34.2 \pm 0.0$   & $70.4 \pm 2.0$              \\
 medium-replay           & halfcheetah            & $55.1 \pm 1.0$              & $53.0 \pm 6.1$   & $41.0 \pm 3.8$              \\
 medium-replay           & hopper                 & $89.5 \pm 1.8$              & $41.3 \pm 11.3$  & $34.5 \pm 19.1$             \\
 \textbf{medium-replay}           & \textbf{walker2d}               & $56.0 \pm 8.6$              & $32.1 \pm 25.7$  & $\mathbf{70.9 \pm 4.3}$              \\
 med-expert           & halfcheetah            & $90.0 \pm 5.6$              & $37.6 \pm 0.0$   & $86.4 \pm 3.5$              \\
 med-expert           & hopper                 & $111.1 \pm 2.9$             & $34.6 \pm 30.4$  & $18.7 \pm 4.6$              \\
 med-expert           & walker2d               & $103.3 \pm 5.6$             & $48.3 \pm 68.0$  & $54.2 \pm 25.5$             \\
\hline
\end{tabular}
}
\end{center}
\label{tab:mujoco_results}
\end{table*}

\begin{figure}[t]
\centering
    \resizebox{0.95\linewidth}{!}{%
        \includegraphics[width=0.98\textwidth]{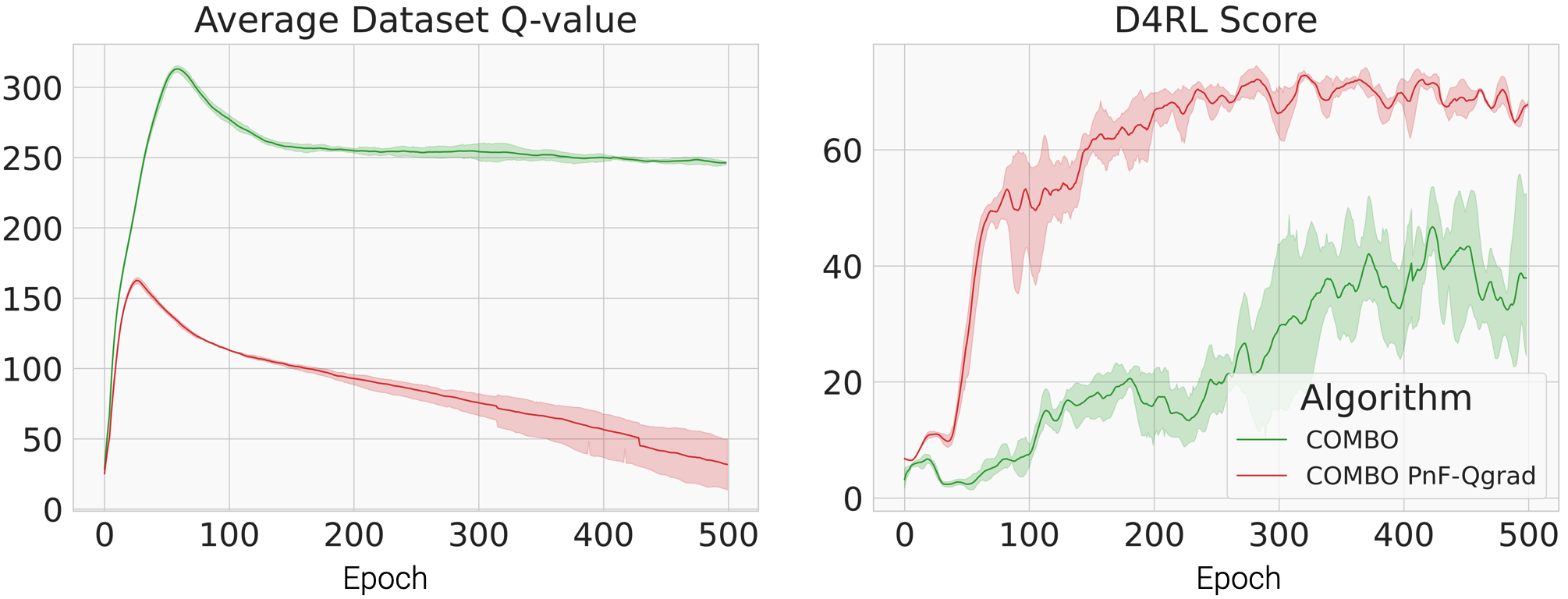}
    }
\caption{%
Average dataset Q-value (left) and D4RL score (right) for the 
\texttt{Walker2d-v2 Medium-Replay} task over 6 random seeds.
The training curve is for 500 epochs of 
the policy update phase that occurs after an initial phase of 
fitting a model to the offline dataset. A consistently and significantly 
lower overall average dataset Q-value is obtained for \texttt{COMBO \PNF}
in comparison to the \texttt{COMBO} baseline. 
COMBO \protect\citep{yu2021combo} advocates for lower average dataset Q-values
as they prevent overestimation of Q-value for unseen states and actions, while
also being strongly linked to higher online evaluation performance (D4RL score).
We observe a similar result i.e. 
lower average dataset Q-values for \texttt{COMBO \PNF} lead to better D4RL score.
}
\label{fig:adq_fig}
\end{figure}

In \Cref{fig:dset_frac}, we also demonstrate the effect of reducing
dataset size for the \texttt{Walker2D-v2 Medium-Replay} task. We
reduce dataset size by selecting a contiguous sub-arrays 
from the front of the original dataset given a fraction (e.g. fraction
$0.5$ takes the first half of the dataset). While fractions of $0.5$
and smaller rapidly reduce performance and make them comparable across methods,
we observe that the average dataset Q-value remains significantly lower for
\texttt{COMBO \PNF} in comparison to \texttt{COMBO}.

\begin{figure}[t]
\centering
    \resizebox{1.0\linewidth}{!}{%
        \includegraphics[width=0.98\textwidth]{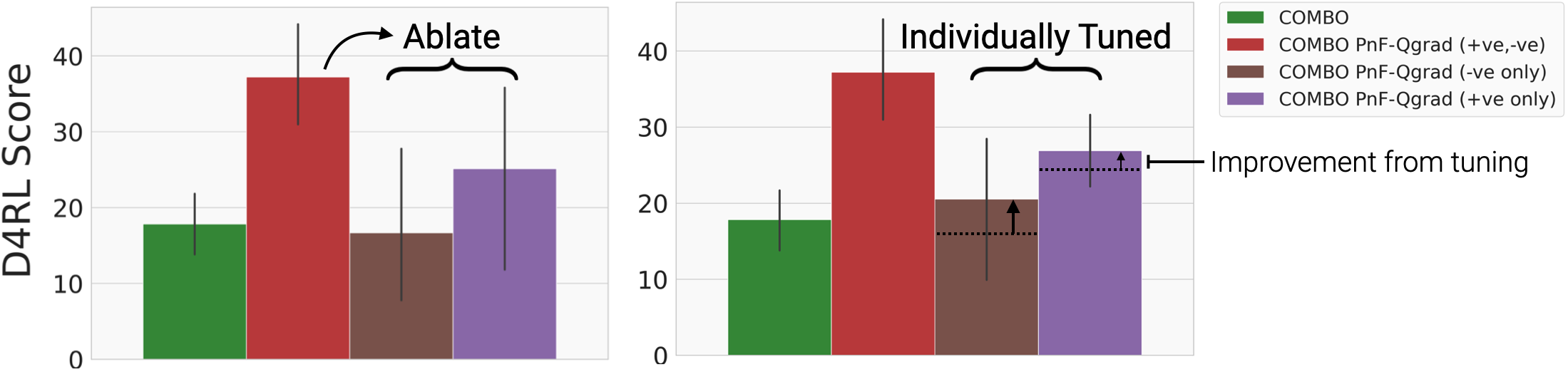}
    }
\caption{%
D4RL Score comparison on \texttt{Pen-v1 Human} task 
over 6 random seeds of ablations of \PNF that use 
positive-only step size along Q-gradient
(i.e. $\eta_{s} \sim U(0, \delmax)$ in \Cref{alg:value_aware_augment}),
and negative-only step size along Q-gradient i.e. 
(i.e. $\eta_{s} \sim U(-\delmax, 0)$ in \Cref{alg:value_aware_augment}).
(left) Ablations for positive-only and negative-only inherit the 
hyperparameter values ($\nsteps, \faug, \delmax$) from 
\PNF (positive, negative) (red bar). (right) Each ablation is individually
tuned to obtain best value of hyperparameters $\nsteps, \faug, \delmax$.
}
\label{fig:posneg}
\end{figure}

\begin{figure*}[t]
\centering
    \resizebox{0.9\linewidth}{!}{%
        \includegraphics[width=0.98\textwidth]{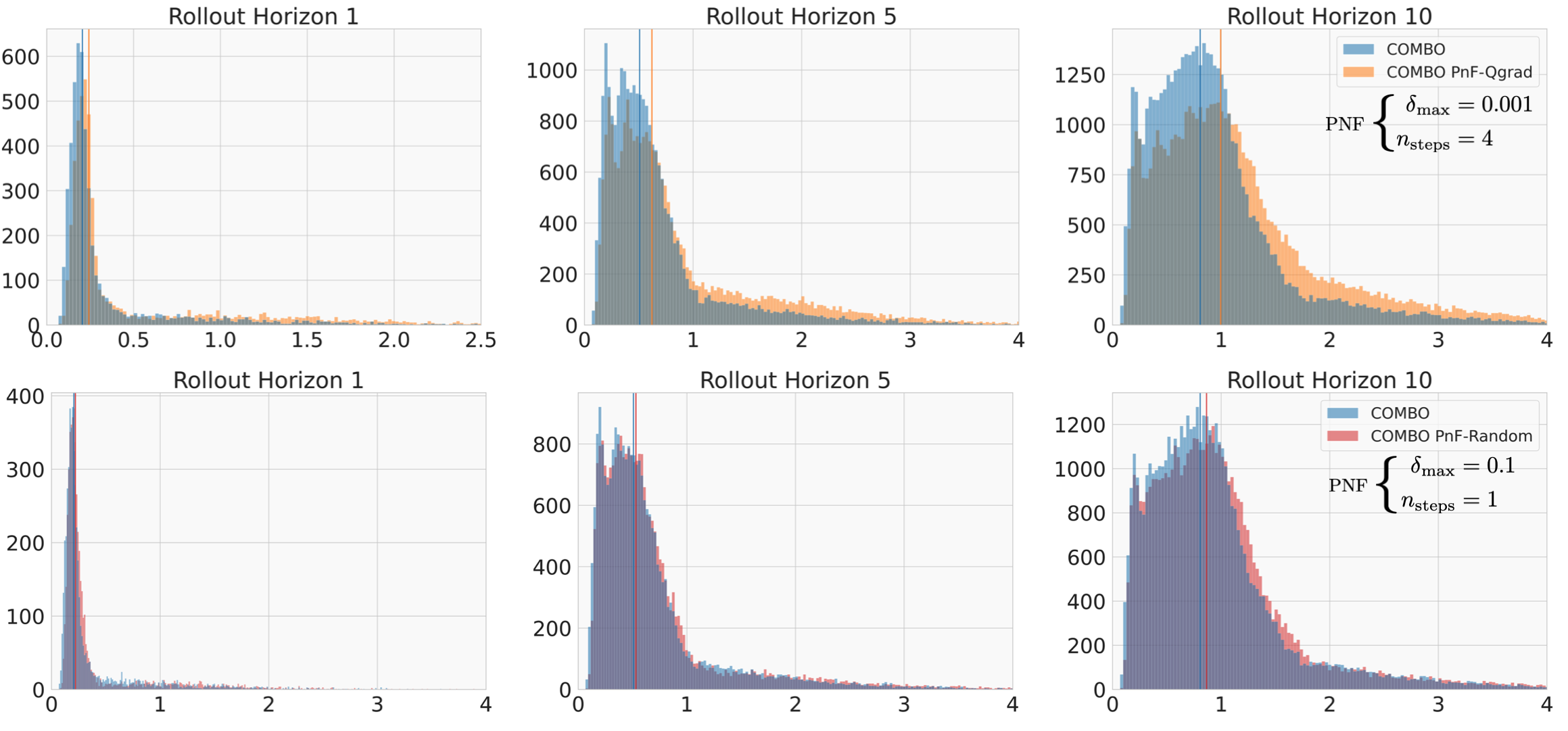}
    }
\caption{%
Histograms of distance of visited unseen states from seen dataset computed using 
L2 distance from nearest neighbor in seen dataset (X-axis) for the 
\texttt{Pen-v1 Human} dataset. 
Unseen states in the COMBO
baseline are obtained by model rollouts from seen states, 
whereas unseen states
from \PNF and \PNFRand are obtained using model rollouts starting from 
augmented states using \Cref{alg:value_aware_augment}
and their respective choice of perturbation directions. 
Rollout horizon values are (left to right) $1, 5, 10$.
\PNF uses $\delmax = 0.001$, $\nsteps=4$ and \PNFRand uses 
$\delmax=0.1$, $\nsteps=1$, which were the best hyperparameter
values selecting after tuning. Dark colored 
full length vertical lines correspond
to median of respective distribution.
}
\label{fig:l2_dist_analysis}
\end{figure*}

\subsection{Sign of Perturbation}
In \Cref{alg:value_aware_augment}, we choose the step size 
$\eta_{s}\in U(-\delmax, \delmax)$ for a given direction 
$\nabla_{s} Q_{\theta}(s, \pi_{\theta}(s))$. In order to verify that
both positive and negative step sizes for perturbation along this direction
are important for good performance, we evaluate two ablations
that use either a positive-only step size or a negative-only step size,
using the \texttt{Pen-v1 Human} task. We refer to 
\texttt{COMBO \PNF (+ve, -ve)} as our original method that samples 
$\eta_{s}\in U(-\delmax, \delmax)$, \texttt{COMBO \PNF (+ve-only)}
as the positive-only ablation that samples 
$\eta_{s}\in U(0, \delmax)$ and 
\texttt{COMBO \PNF (+ve-only)} as the negative-only ablation that samples
$\eta_{s}\in U(-\delmax, 0)$. \Cref{fig:posneg} highlights our findings
in two settings. First, we use inherit the hyperparameter values
for each ablation from the \texttt{COMBO \PNF (+ve, -ve)}. In this setting,
we find that performance is immediately lowered for both ablations,
indicating that either both positive and negative step sizes matter,
or that the ablations may need to be individually tuned. Second,
we eliminate one of the possibilities by individually tuning
hyperparameter values for both ablations, where performance is slightly
improved over the un-tuned methods, while still having 
lower performance than \texttt{COMBO \PNF (+ve, -ve)}. As a result,
we find that it is important to perturb along the positive as well as
negative direction of Q-value gradient for best performance.

\subsection{Distribution of Perturbation Distance}
In order to understand the impact of our studied perturbations
on the distance of augmented unseen states from seen states,
we plot a histogram of nearest neighbor distances of every
unseen state (w.r.t seen states in the entire offline dataset) 
visited by \Cref{alg:mboffrl_ours} (for \PNF and \PNFRand)
and \Cref{alg:mboffrl_combo} (COMBO). For each algorithm,
the set of unseen states are an aggregation of all states visited
during model rollouts. For \Cref{alg:mboffrl_ours}, we set the fraction of 
model rollouts that occur from unseen states, $\faug$, to $1.0$
We expect that any perturbation based unseen state augmentation 
has the potential to visit states in model rollouts that are far 
away from any nearest seen state. 

\Cref{fig:l2_dist_analysis} 
plots these histograms for \PNF, \PNFRand and COMBO on the 
\texttt{Pen-v1 Human} task with rollout horizon values
in $\{1, 5, 10\}$. For \PNF, we set $\delmax=0.001, \nsteps=4$,
which are the tuned values of these hyperparameters for this task.
Similarly, for \PNFRand, we set $\delmax=0.1, \nsteps=1$ (note that
for \PNFRand, $\nsteps$ is always set to $1$ as multiple gradient steps
are not necessary for a random perturbation direction). We find that
despite the low value of $\delmax$, \PNF is able to find unseen state
augmentations that produce model rollout states with 
farther nearest neighbor L2 distance than the 
model rollouts states visited by no augmentation (COMBO baseline).
This is in comparison to \PNFRand, which has overall distribution
of distances similar to no augmentation (despite the higher $\delmax$).
This indicates that \PNF is able to find unseen states farther from
seen data while still passing through uncertainty filter, whereas
random perturbation are not able to do the same.

\section{Conclusion}
In this paper, we addressed the limitations of model-based offline reinforcement learning methods
in their ability to find and utilize unseen states for Q-value estimation. 
We proposed an unseen state augmentation strategy that is able to find states that are 
far away from the seen states in the offline dataset, have low estimated epistemic 
uncertainty and which lead to overall lower Q-value estimates and better performance in 
several offline RL tasks. We present a value-informed perturbation strategy that uses the
positive and negative Q-value gradient direction to generate unseen state proposals; and find
improved performance compared to an ablation that perturbs uniformly randomly in all directions
and ablations that use the positive or negative Q-value gradient direction alone.

\bibliography{refs}
\bibliographystyle{icml2023}

\newpage
\appendix
\label{app: offrl_hyperparam}
\onecolumn

\end{document}